\documentclass{article}

\usepackage[final]{neurips_2022}

\usepackage[utf8]{inputenc} 
\usepackage[T1]{fontenc}    
\usepackage{hyperref}       
\usepackage{url}            
\usepackage{booktabs}       
\usepackage{amsfonts}       
\usepackage{nicefrac}       
\usepackage{microtype}      
\usepackage{xcolor}         
\usepackage{physics}
\newcommand{\ours}[0]{\textsc{Qwale}\xspace}
\usepackage{wrapfig}

\title{You Only Live Once: \\
Single-Life Reinforcement Learning} 

\author{%
  Annie S. Chen$^1$, Archit Sharma$^1$, Sergey Levine$^2$, Chelsea Finn$^1$ \\
  Stanford University$^1$, UC Berkeley$^2$\\
  \texttt{asc8@stanford.edu} \\
}

\usepackage{algorithm}
\usepackage[noend]{algorithmic}
\usepackage{wrapfig}
\usepackage{amsmath}
\usepackage{amssymb}
\usepackage{mathtools}
\usepackage{subcaption}
\usepackage{ragged2e}
\usepackage{booktabs}
\usepackage{tikz}
\usepackage{etoolbox}
\usepackage{xspace}
\usepackage{dsfont}
\usepackage{xpatch}
\usepackage{enumerate}
\usepackage{xstring}
\usepackage{bbm}
\usepackage{amsthm}
\usepackage{setspace}
\usepackage{tabularx}
\usepackage{graphicx}

\newcommand{\algo}{\textsc{Qwale}\xspace}
\newcommand{\slrl}{SLRL\xspace}

\newcommand{\mc}{\mathcal}
\definecolor{mygreen}{rgb}{0.1, 0.6, 0.1}

\newcommand{\E}{\mathbb{E}}

\newcommand{\Dprior}{\mathcal{D}_{\text{prior}}}

\usepackage{titlesec}
\titlespacing\section{0pt}{0pt plus 2pt minus 2pt}{0pt plus 2pt minus 2pt}
\titlespacing\subsection{0pt}{3pt plus 4pt minus 2pt}{0pt plus 2pt minus 2pt}
\titlespacing\subsubsection{0pt}{3pt plus 4pt minus 2pt}{0pt plus 2pt minus 2pt}

\begin{document}

\maketitle


\begin{abstract}

Reinforcement learning algorithms are typically designed to learn a performant policy that can repeatedly and autonomously complete a task, usually starting from scratch. However, in many real-world situations, the goal might not be to learn a policy that can do the task repeatedly, but simply to perform a new task successfully once in a single trial. For example, imagine a disaster relief robot tasked with retrieving an item from a fallen building, where it cannot get direct supervision from humans. It must retrieve this  object within one test-time trial, and must do so while tackling unknown obstacles, though it may leverage knowledge it has of the building before the disaster. We formalize this problem setting, which we call \emph{single-life reinforcement learning} (SLRL), where an agent must complete a task within a single episode without interventions, utilizing its prior experience while contending with some form of novelty. 
SLRL provides a natural setting to study the challenge of autonomously adapting to unfamiliar situations, and we find that algorithms designed for standard episodic reinforcement learning often struggle to recover from out-of-distribution states in this setting.
Motivated by this observation, we propose an algorithm, $Q$-weighted adversarial learning (\textsc{Qwale}), which employs a distribution matching strategy that leverages the agent's prior experience as guidance in novel situations. Our experiments on several single-life continuous control problems indicate that methods based on our distribution matching formulation are 20-60\% more successful because they can more quickly recover from novel states. \footnote{\small Project website: \url{https://sites.google.com/stanford.edu/single-life-rl}}

\end{abstract}

\section{Introduction}

Agents operating in the real world must often contend with novel situations that differ from their prior experience. For example, a search-and-rescue disaster relief robot may encounter novel obstacles while traversing a building. The vast range of situations that could be encountered at test time cannot be anticipated in advance during training, and therefore agents that can adapt at test time will be better equipped to cope with new situations. Reinforcement learning (RL) algorithms in principle provide a suitable paradigm for such adaptation. However, much of the work in RL focuses on learning an optimal policy that can repeatedly solve a given task, whereas on-the-fly adaptation simply requires the agent to complete the task once without human interventions or supervision, presenting a different set of challenges. In this work, we formalize this problem setting as \textit{single-life reinforcement learning} (SLRL), where the agent is evaluated on its ability to successfully and autonomously adapt to a novel scenario within a single trial.

\begin{figure*}[t!]
    \centering
    \includegraphics[width=\textwidth]{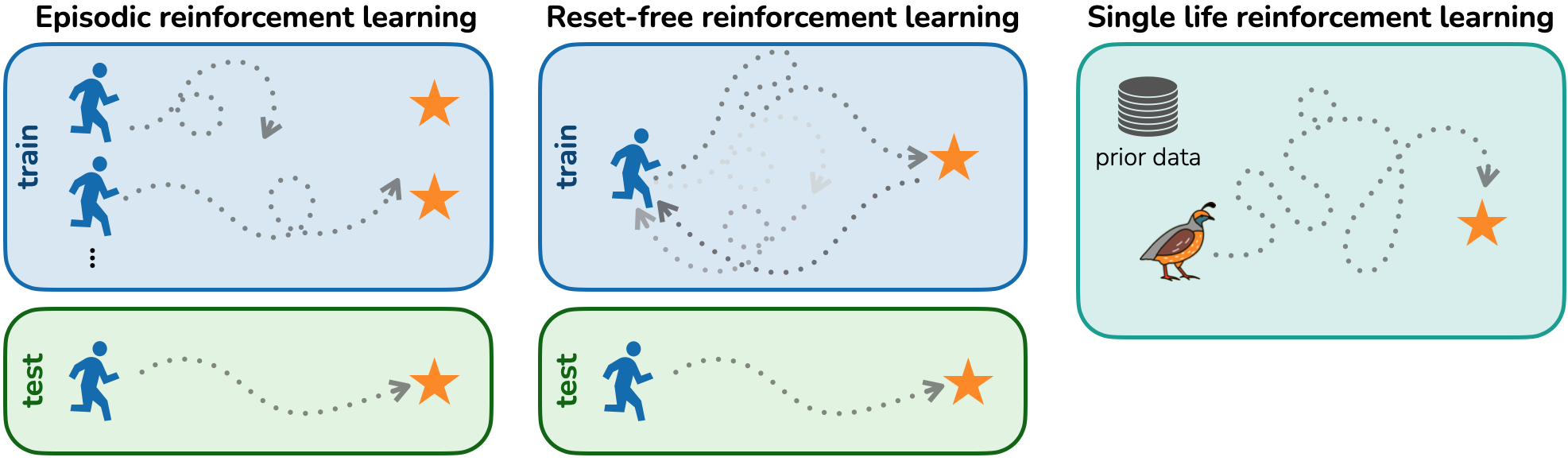}
    \vspace{-0.4cm}
    \caption{
        \small We study the single-life reinforcement learning (\slrl) problem, where given prior data, an agent must complete a task autonomously in a single trial in a domain with a novel distribution shift.
    }
    \vspace{-2mm}
    \label{fig:teaser}
\end{figure*}

Given some prior experience, \slrl induces the challenge of contending with novelty (e.g., a new initial state distribution or new dynamics) at test time without episodic resets. A key problem that makes this difficult is that the agent will need to recover from unfamiliar situations without any interventions, which can be especially challenging when informative shaped rewards are unavailable, as is often the case in the real world. In fact, task rewards are sometimes only provided at the end of the life, e.g. the search-and-rescue robot may only receive reward signal after it maneuvers to find a desired object and may therefore get stuck behind unknown debris. 
In episodic RL, the agent can rely on a reset to recover from an unfamiliar state. In contrast, in \slrl, the agent only has its prior experience to leverage in order to find its way back to a good state distribution. Consequently, we find that fine-tuning a pre-trained value function via online RL will struggle in \slrl settings, as it does not explicitly encourage the agent to return to its known distribution and complete the desired task.


To better handle the \slrl setting, we need a way to provide intermediate shaped rewards that guide the adaptation process, and help the agent recover when it gets stuck. One potential option for providing the desired guidance is to use reward shaping towards the agent's distribution of prior experience. Adversarial imitation learning (AIL) approaches such as GAIL (\cite{ho2016generative}) can potentially provide such reward shaping, but using existing AIL methods naively may not give the intended behavior in the \slrl setting due to two main shortcomings. First, such methods assume expert demonstrations are given as prior data, but in \slrl, we may be given suboptimal offline prior data. Second, AIL methods train the agent to match the entire distribution of prior data, which may be key to learning an optimal policy, but may be a drawback in our setting, as the agent might not be consistently guided towards task completion. To address these shortcomings, we propose $Q$-weighted adversarial learning (\algo), in which different states in the prior data are weighted different amounts by their estimated $Q$-value. \algo incentivizes the agent to move towards states in the prior data with higher values, so that agent is more consistently guided towards states closer to task completion.


Our contributions are as follows. We formalize the \slrl problem setting, which provides a natural setting in which to study autonomous adaptation to novel situations in reinforcement learning. Moreover, we believe \slrl is a useful framework for modeling many situations in the real world.
We explore reward shaping through distribution matching as one potential choice of methods for \slrl and identify challenges that uniquely arise in the \slrl setting with existing distribution matching approaches. We propose a new approach, $Q$-weighted adversarial learning (\algo), which is less sensitive to the quality of prior data available and provides the agent with a shaped reward towards completing the desired task a single time. Through our experiments, we find that \algo can meaningfully guide the agent to recover to the prior data distribution to complete the desired task 20-60\% more successfully on four separate domains compared to existing distribution matching approaches and RL fine-tuning. \algo provides a strong baseline for developing algorithms that can better adapt to novelty online and recover from out-of-distribution states.



\section{Related Work}
\label{sec:related}

\textbf{Autonomous RL}. In the context of deep RL, agents typically (but not always) are trained in episodic setting and are evaluated on the quality of the learned policy. Several recent works have developed algorithms that can learn without episodic resets~\citep{han2015learning, eysenbach2017leave, zhu2020ingredients, sharma2021subgoal, sharma2022state, gupta2021reset,gupta2022bootstrapped, kim2022automating,sharma2021autonomous}. Like our work, such methods aim to make it possible to learn without any episodic resets, but are typically still focused on acquiring an effective policy that can perform the task repeatedly, typically by training some auxiliary controller to enable the policy to ``retry'' the task multiple times without resets. In contrast, our aim is to develop an algorithm that can solve the task once, but as quickly as possible, which introduces a unique set of challenges as we discussed above.

\textbf{Continual RL}. There is a rich literature on reinforcement learning in the continuing setting~\citep{mahadevan1996average, schwartz1993reinforcement, sutton2018reinforcement, xu2018reinforced, NEURIPS2019_fa7cdfad, Lomonaco_2020_CVPR_Workshops, pmlr-v119-wei20c} that considers maximizing the average reward accumulated over an infinite horizon without episodic resets. Such works often also consider regret minimization as the objective. SLRL is closely related, and can be viewed as a special case where the agent has access to a prior offline dataset and aims to solve a single task as quickly as possible in a new domain with a distribution shift. While the focus in continual learning is on general ``lifelong'' methods or on exploration, our focus is on effectively leveraging prior data in a setting that is meant to be reflective of real-world tasks (for example, in robotics).

\textbf{Leveraging offline data in online RL}. Learning expert policies given prior interaction data has been extensively studied in imitation learning~\citep{argall2009survey, ross2011reduction, ghasemipour2020divergence}, inverse RL~\citep{ng2000algorithms,finn2016guided, ziebart2008maximum, ziebart2010modeling}, RL for sparse reward settings~\citep{brys2015reinforcement, nair2018overcoming, rajeswaran2017learning, hester2018deep, vecerik2017leveraging} and offline RL~\citep{wang2020critic, levine2020offline, kumar2020conservative, kidambi2020morel, wu2019behavior, nair2018overcoming}. Across all these diverse topics, the goal is to learn a competent policy that can solve the task efficiently whereas the objective in this work is to complete the task in a single trial as quickly as possible. To this end, we build on recent adversarial approaches to inverse RL~\citep{ho2016generative, fu2017learning, singh2019end, torabi2019adversarial, kostrikov2018discriminator, zhu2020off} to encourage agent's state visitation towards expert prior data where the agent is likely to be successful. Prior methods have also studied adversarial inverse RL and imitation learning with non-expert data~\citep{wang2018exponentially, sun2019adversarial, wu2019imitation, wang2020support, wang2021learning, wang2021robust, cao2022learning, beliaev2022imitation}. However, as we discuss in Section~\ref{sec:rebuttal-method}, these approaches need to be adapted for the \slrl setting to be efficient at completing the task and to handle novelty, for example when the dynamics may have changed. 

\textbf{Transfer and adaptation in RL}. Many prior works have studied the problem of adapting in presence of shifts between train and test settings, often in a specific problem setting such as sim2real transfer~\citep{sadeghi2016cad2rl,tobin2017domain, peng2018sim, mehta2020active} or fast adaptation via meta-learning~\cite{finn2017model, nichol2018first, nagabandi2018learning, zintgraf2019varibad, finn2019online}. A common theme in these works is that the algorithm can often train in preparation for adaptation at test-time, thus affecting the prior experiences it may collect. In contrast, the \slrl setting lays algorithmic emphasis on online exploration and adaptation, as the agent has access to a fixed prior dataset of experiences. Other transfer learning approaches adapt the weights of the policy to a new environment or task, either through rapid zero-shot adaptation~\cite{hansen2020self, yoneda2021invariance} or through extended episodic online training~\cite{khetarpal2020towards,rusu2016progressive,eysenbach2020off,xie2020deep, xie2021lifelong}. Unlike the latter, we focus on adaptation within a single episode, but, unlike the former, with a focus on extended exploration and learning over tens of thousands of timesteps. This problem setting leads to unique challenges, namely that the agent must autonomously recover from mistakes, hence requiring a distinct approach.

\section{Preliminaries}
\label{sec:prelim}

In this section, we describe some preliminaries before formalizing our problem statement in the following section. We consider an agent that operates in a Markov decision process (MDP) consisting of the tuple $\mc{M} = (\mc{S}, \mc{A}, \mc{P}, \mc{R}, \rho_0, \gamma)$, where $\mc{S}$ is the state space, $\mc{A}$ is the agent's action space, $\mc{P}(s_{t+1} | s_t, a_t)$ represents the environment's transition dynamics, $\mc{R}: \mc{S} \rightarrow \mathbb{R}$ indicates the reward function, $\rho_0: \mc{S} \rightarrow \mathbb{R}$ denotes the initial state distribution, and $\gamma \in [0, 1)$ denotes the discount factor. In typical reinforcement learning, the objective is find a policy $\pi$ that maximizes $J(\pi) = \E_{\tau \sim \pi}[\sum_{t=0}^{\infty} \gamma^t \mc{R}(s_t)]$. 

Although our method is a reward-driven RL algorithm, we utilize concepts from imitation learning to overcome sparse rewards, utilizing potentially suboptimal prior data. To this end, we build on adversarial imitation learning (AIL), which uses prior data $\Dprior$ in the form of expert demonstrations (we will relax this requirement) to recover the expert's policy. 

One such method is GAIL~\citep{ho2016generative}, which finds a policy $\pi_{\theta}$ that minimizes the Jensen-Shannon divergence between its stationary distribution and the expert data. It does so by training a discriminator network $D: \mc{S} \times \mc{A} \rightarrow (0, 1)$, alternating updates with updates to the policy $\pi$. Concretely, $D$ and $\pi$ are learned by optimizing the following:
\begin{align}
    &\min_{\pi} \max_{D \in (0, 1)^{\mc{S} \times \mc{A}}} \E_{\pi_E} [\log(D(s, a))] + \E_{\pi}[\log (1 - D(s, a))] - \lambda \mathcal{H}(\pi) \nonumber \\
    &= \min_{\pi} \mathcal{D}_{\textrm{JS}}(\rho^\pi(s, a) \mid\mid \rho^*(s, a)) - \lambda \mathcal{H}(\pi),
    \label{eq:gail}
\end{align}
where $\mathcal{D}_{\textrm{JS}}$ is the Jensen-Shannon divergence and $\rho^*(s, a)$ is the optimal state-action occupancy measure of the expert $\pi_E$ (represented by demonstrations).
In Section~\ref{sec:rebuttal-exps}, we will see that AIL-style discriminator-based approaches can be adapted to the \slrl problem setting \emph{without} demonstration data, and will therefore form the basis of our method.

\section{Single-Life Reinforcement Learning}
\label{sec:problem}

In this section, we formalize the \emph{single-life reinforcement learning (\slrl)} problem. The defining characteristic of \slrl is that the agent is given a single ``life'' (i.e., one long trial) to complete a desired task, with the trial ending when the task is completed. The agent must complete the task autonomously, without access to any human interventions or resets. Furthermore, as is the case in many real-world situations, when faced with situations where a task must be completed once, informative shaped rewards are not easily obtained, and so the single life might have a reward only once after the task has been completed. 
However, an agent typically has some prior knowledge. E.g., an agent tasked with retrieving a valuable from a burning building may have experience finding items in that building before the fire. We will therefore assume access to offline prior data of some sort that the agent may use for pretraining and during its single life.
In many cases such as the disaster relief example, we may not have expert prior data of the desired task in the desired environment. 
Hence, we emulate this setting by providing the agent with prior data from a related environment and deploying its single life on a domain with a distribution shift. 


We can formalize this setting as follows. We are given prior data $\Dprior$, which consists of transitions from some source MDP $\mc{M}_{\text{source}}$. The agent will then interact with a target MDP defined by $\mc{M}_{\text{target}} = (\mc{S}, \mc{A}, \Tilde{\mathcal{P}}, \mc{R}, \Tilde{\rho}_0, \gamma)$. We assume that the the target MDP has an aspect of novelty not present in the source MDP, such as different dynamics $\Tilde{\mathcal{P}}(s_{t+1} \mid s_t, a_t)$ or a different initial state distribution $\Tilde{\rho}$. Naturally, the more similar the domains are, the easier the problem becomes, and the effectiveness of any algorithm will be strongly dependent on the degree of similarity, though formalizing a precise assumption on similarity between the source and target domain is difficult. The reward between the two MDPs is the same, meaning the agent is still trying to accomplish the same task in the target domain as in the source. 

Maintaining the same notation as the previous section, the \slrl problem aims to maximize $J = \sum_{t=0}^h \gamma^t \mc{R}(s_t)$, where $h$ is the trial horizon, which may be $\infty$. In general, we expect the task reward to be such that learning \emph{only} from task rewards during the single life is difficult, as the reward may be very sparse or even awarded only upon successful completion of the task. We assume that there are no sink states beyond a terminal success state, such that it is possible for the agent to autonomously recover from any mistake.
Note that this setup is essentially the same as the widely studied regret minimization problem in exploration~\citep{mahadevan1996average}. However, while regret minimization is typically studied in the context of RL exploration theory, our aim with \slrl is to study a particular \emph{special case} of the more general regret minimization framework that is meant to reflect a realistic setting in real-world RL (e.g., robotics) where an agent with prior experience must solve a task in a single (potentially long) trial.


As we analyze in Section~\ref{sec:rebuttal-exps}, algorithms designed for episodic policy learning do not perform well in the \slrl problem setting, even when the policy and replay buffer are pre-trained and seeded with the prior data, because they do not quickly recover from mistakes to get back onto a good state distribution. In Section~\ref{sec:rebuttal-method}, we will discuss an approach based on distribution matching that attempts to address this issue.
\section{$Q$-weighted Adversarial Learning (\algo)}
\label{sec:rebuttal-method}

In this section, we will present our method for addressing \slrl, which we call \algo. The key insight in \algo is to utilize the prior data $\Dprior$ to efficiently complete the task, especially when the reward information is sparse in the target domain. 
The framework of AIL provides a reasonable starting point, though it is not sufficient by itself: existing AIL methods assume that the prior data consists of expert demonstrations, and they train the agent to match the entire demo distribution.
Since our goal is not to learn a policy that repeatedly performs the task, but rather to solve it as quickly as possible once, we do not actually want to \emph{learn to imitate} prior data, but rather to seek out states that resemble the \emph{best} states in the prior data, with better states being more preferred. This is especially important when the prior data is not actually optimal, but might consist of arbitrarily suboptimal states. We will discuss how this can be accomplished with a modification of AIL which, instead of treating prior data as equally desirable, preferentially drives the agent toward states that resemble the best states in the prior data.

\subsection{Algorithm Description}
Our algorithm’s desired behavior is to lead the agent towards the distribution of prior data, therefore helping it recover when it falls into out-of-distribution states, while also pushing it towards task completion. GAIL~\citep{ho2016generative} offers a framework to shape the state-action distribution towards that of an expert. In particular, GAIL minimizes $\mathcal{D}_{\textrm{JS}}(\rho^\pi(s, a) \mid\mid \rho^*(s, a))$ (ignoring the causal entropy $\mathcal{H}(\pi)$). In \slrl, we relax this assumption, as we may only have access to suboptimal offline prior data. As such, we desire an algorithm that will be agnostic to the quality of prior data. Hence, instead of learning a policy that aims to uniformly match the entire state-action distribution of the prior data, we want to match the state-action pairs that will lead to task completion. 

The problem of matching the state-action distribution has been studied in stochastic optimal control, particularly in REPS~\citep{peters2010relative}. Roughly, this problem can be framed as the RL problem with an additional constraint encouraging the learned distribution to be close to the distribution of the prior data.
The solution to this optimization problem gives us that our desired target state-action distribution has the form $\rho^*_{\textrm{target}}(s, a) \propto \rho_\beta(s, a) \exp (Q^{\pi_{\textrm{target}}}(s, a) - V^{\pi_{\textrm{target}}}(s))$, where the corresponding policy can be derived by using $\pi_{\textrm{target}} = \rho^*_{\textrm{target}}(s, a)/ \int_\mathcal{A}\rho^*_{\textrm{target}}(s, a') \dd a'$ using~\citep{syed2008apprenticeship}. For a formal description of the problem, see Appendix A.1.

Thus, replacing $\rho^*_{\textrm{target}}$ as the target distribution instead of $\rho^*$, we propose $Q$-weighted adversarial learning (\algo), which minimizes $\mathcal{D}_{\textrm{JS}}(\rho^\pi \mid\mid \rho^*_{\textrm{target}})$ as follows:
\begin{align*}
    \min_\pi &\mathcal{D}_{\textrm{JS}}(\rho^\pi(s, a) \mid\mid \rho^*_{\textrm{target}}(s, a)) = \min_\pi \max_D \mathbb{E}_{s, a \sim \rho^*_{\textrm{target}}} \left[\log D(s, a)\right] + \mathbb{E}_{s, a \sim \rho^\pi}\left[\log(1 - D(s, a))\right]\\
    &= \min_\pi \max_D \mathop{\mathbb{E}}_{s, a \sim \rho^\beta} \left[\frac{\exp(Q^{\pi_{\textrm{target}}}(s,a) - V^{\pi_{\textrm{target}}}(s))}{\mathbb{E}_{\rho^\beta}\left[\exp(Q^{\pi_{\textrm{target}}}(s,a) - V^{\pi_{\textrm{target}}}(s))\right]}\log D(s, a)\right] + \mathop{\mathbb{E}}_{s, a \sim \rho^\pi}\left[\log(1 - D(s, a))\right]\\
    &\equiv \min_\pi \max_C \mathbb{E}_{s, a \sim \rho^\beta} \left[\exp\left(Q^{\pi_{\textrm{target}}}(s,a) - V^{\pi_{\textrm{target}}}(s)\right)\log D(s, a)\right] + \mathbb{E}_{s, a \sim \rho^\pi}\left[\log(1 - D(s, a))\right],
\end{align*}
where the constant $\mathbb{E}_{\rho^\beta}\left[\exp(Q^{\pi_{\textrm{target}}}(s,a) - V^{\pi_{\textrm{target}}}(s))\right]$ can be ignored.

Hence, \algo trains a weighted discriminator using a fixed $Q$-function $Q(s, a)$ trained in the source MDP to distinguish between useful transitions and ones that may be less useful. This $Q$-function may be obtained through RL pretraining, which is what we use for our experiments, or a variety of other ways, such as offline RL or Monte Carlo estimation. We train the discriminator in a similar manner as GAIL, where the positives come from offline data and negatives from online experience. 

When training the discriminator, we weight the positive states $s$ by $\exp(Q(s, a) - b)$, where $b$ is an implementation detail treated as a hyperparameter. Following~\cite{abdolmaleki2018maximum}, we exclude $V^{\pi_{\textrm{target}}}(s)$ in the weighting term in our practical implementation. Including the term $b$ changes the bias on the discriminator, which in effect just adds a constant to the reward. In practice, to avoid having to tune $b$, we just use the value of the most recent state as $b$, i.e. $b = Q(s_t, a_t)$. We include additional discussion in the Appendix. 
We normalize the $Q$-values to be between 0 and 1 by subtracting the min and dividing by the (max - min) and train the $Q$-weighted discriminator in alternating updates that finetune the policy and critic in an AIL fashion. In this manner, \algo extends AIL to the general setting with any prior data. 

Intuitively, the goal of our weighted discriminator training procedure is to obtain a discriminator that, when used as a reward, will drive the agent toward states that it believes would lead to better outcomes than its present state, based on the prior data. The $Q$-function quantifies the agent's belief from the prior data that a particular state will lead to high reward, making this a natural choice for estimating how desirable a state is at any given time.
Hence, using $Q$-values to weight the examples for the discriminator will cause the discriminator to prefer states that are closer to the goal over states that are further away. This is significantly different from the behavior we would expect to see if we were simply imitating the optimal policy, which would give equal weight to all of the transitions along an optimal path.

\subsection{Practical Implementation}
\begin{algorithm}[ht!]
\caption{\small \textsc{Q-Weighted Adversarial Learning (\algo)}}
\begin{algorithmic}[1]
\footnotesize
\STATE \textcolor{blue}{// Single Trial Deployment}
\STATE \textbf{Require:} $\Dprior$, test MDP $\mc{M}_{\text{test}}$, pretrained critic $Q(s, a)$, and (optionally) policy $\pi$; 
\STATE \textbf{Initialize:} replay buffer for online transitions $\mc{D}_{\text{online}}$; parameters $\phi$ for discriminator $q_{\phi}(\text{prior} \mid s_t)$, timestep $t=0$
\WHILE{\text{task not complete}}
    \STATE Take $a \sim \pi( \cdot \mid s_t)$, and observe $r_t$ and $s_{t+1}$
    \STATE $\mc{D}_{\text{online}} \leftarrow \mc{D}_{\text{online}} \cup \{(s_t, a_t, r_t, s_{t+1})\}$ 
    \STATE $\phi \leftarrow \phi - \eta \nabla_{\phi} L(\phi)$ \textcolor{blue}{// Update discriminator according to Eq.~\ref{eqn:disc}}
    \STATE $r'(s_t) = r(s_t) - \log(1-q_{\phi}(\text{prior} \mid s_t))$ 
    \STATE $Q(s, a), \pi \leftarrow \textsc{SAC}(Q(s, a), \pi, \Dprior \cup \mc{D}_{\text{online}}, r')$
    \STATE Increment $t$
\ENDWHILE{}
\end{algorithmic}
\label{algoblock}
\end{algorithm}

We optimize our objective using maximum entropy off-policy RL with the SAC algorithm (\cite{haarnoja2018soft}), modified in a similar manner as we did with GAIL in the previous section. In particular, we learn an additional discriminator $q_{\phi}(\text{prior} \mid s_t)$, optimized using standard cross-entropy loss, which is weighted accordingly.
\begin{equation}
\label{eqn:disc}
     L(\phi) = -\E_{\Dprior} [\exp(Q(s, a) - b)\log q_{\phi}(\text{prior} \mid s)] - \E_{\mc{D}_\text{online}} [\log q_{\phi}(\text{online} \mid s)].
\end{equation}
The discriminator is used to modify the rewards when updating off-policy from all experience--prior and online. At single trial test time, the actor and critic are optionally initialized with the pretrained weights, and the replay buffer is initialized with the offline data. For details such as network architecture and hyperparameters, see Appendix~\ref{sec:appendix}. We present the full algorithm in Algorithm~\ref{algoblock}.

\section{Experiments}
\label{sec:rebuttal-exps}

The goal of our experiments is to answer the following questions: (1) How does \ours compare to prior reinforcement learning and distribution matching approaches in single-life RL settings? (2) Do distribution matching approaches help agents learn to recover from novel situations in single-life RL? (3) How does \ours compare to different variants of adversarial imitation learning, with different prior datasets?

\subsection{Experimental Setup}
To answer the above questions, we construct four single-life RL domains with varying prior datasets and sources of novelty, and then measure performance both in terms of speed of task completion and overall single-life success. In this subsection, we describe this experimental set-up in detail.

\begin{figure*}[t]
    \centering
    \includegraphics[width=0.9\textwidth]{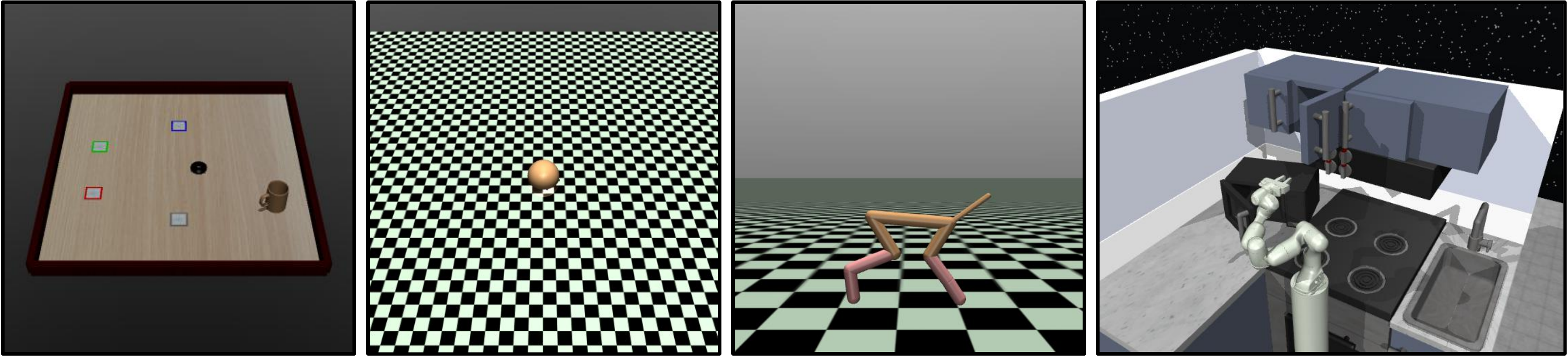}
    \caption{\small
        We evaluate in four different domains, including Tabletop-Organization, Pointmass, HalfCheetah, and a Franka-Kitchen environment with a microwave and cabinet. At test time, an aspect of novelty is introduced in each environment--new initial mug positions for Tabletop, wind for Pointmass, hurdles for the HalfCheetah, and a new combination of tasks for the Franka-Kitchen.
    }
    \label{fig:envs}
    \vspace{-2mm}
\end{figure*}

\textbf{Environments}. We consider the following four problem domains. First, in the Tabletop-Organization environment from the EARL benchmark \cite{sharma2021autonomous}, the agent is tasked with bringing a mug to one of four different locations designated by a goal coaster. The prior data always has the same starting position of the mug. In the target environment, the starting position is in a new location unseen in the prior data. Second, the Pointmass setting tasks an agent to move in 2D from its starting location at the origin $(0, 0)$ to the point $(100, 0)$. The target environment introduces a dynamics shift in the form of a strong ``wind'', where the agent is involuntarily pushed upward in the y-direction each step. Third,  we construct a modified HalfCheetah environment, in which it is difficult but feasible for the cheetah to recover when flipped over. The target environment includes hurdles that the cheetah must jump over, as the prior data does not include these obstacles. Finally, we evaluate on a modified Franka-Kitchen environment, adapted from \cite{gupta2019relay}, where the task is to close a microwave and a hinged cabinet. The prior data only contains trajectories of closing the microwave and the hinged cabinet separately, so the agent must figure out online how to complete both tasks in a row. In other words, both objects are open at the start of single-life RL, and the agent has only previously seen instances where only one is open. For the latter two environments, dense rewards are given, and the discriminator-based reward is added to the extrinsic reward during single-life training. These environments are shown in Figure~\ref{fig:envs}. Further details on the environments are given in Appendix~\ref{sec:appendix}.

\textbf{Comparisons.} To answer question (1), we compare \ours to three alternative methods: (a) SAC fine-tuning, which pre-trains a policy and value function in the source setting and fine-tunes for a single, long episode in the target environment, (b) SAC-RND, which additionally includes an RND exploration bonus~\cite{burda2018exploration} during single-life fine-tuning, and (c) GAIL-s, which runs generative adversarial imitation learning~\cite{ho2016generative, kostrikov2018discriminator} where the discriminator only operates on the current state $s$. We choose for the discriminator to only look at $s$ so that it is less susceptible to dynamics shift between the source data and target environment. We additionally compare to GAIL-sa, which passes both the current state and action to the discriminator. All methods use soft actor-critic (SAC)~\cite{haarnoja2018soft} as the base RL algorithm. We present additional comparisons in the Appendix.

\textbf{Prior datasets}. For all four environments, we evaluate \slrl using data collected through RL as our prior data. More specifically, we run SAC in the source MDP in the standard episodic RL setting for $K$ steps and take the last 50,000 transitions as the prior data. $K$ is chosen such that the prior data contains some good transitions but has not converged to an optimal policy yet. While we are able to run episodic RL in the source MDP, this is not a requirement for \slrl, as long as prior data in the source MDP is available. 
For all methods, including \ours, GAIL variants, and SAC fine-tuning variants, the policy and value function are pretrained in this manner for the initialization of single-life RL. We note that AIL methods like GAIL typically assume that the prior data consists of expert demonstrations but we apply the algorithm only using mixed quality prior data, unless otherwise noted. In particular, Section~\ref{sec:demo-prior} further evaluates AIL methods using demos as prior data, using 10 demonstrations for the Tabletop environment and 3 demonstrations for the Pointmass domain. We include such experiments to answer question (3), i.e. to investigate how the quality of prior data may affect performance.

\textbf{Evaluation Metrics.} To evaluate each method in each environment, we report the average and median number of steps taken before task completion across 10 seeds along with the standard error and success rate (out of 10).
During single-life RL, for all environments, the agent is given a maximum of 200,000 steps to complete the task. If it has not completed the task after 200k steps, then 200k is logged as the total number of steps, and the run is marked as unsuccessful. 

\subsection{Results using mixed data as prior data}

In this subsection, we aim to answer our first experimental question and study how \algo performs compared to prior reinforcement learning and distribution matching approaches in single-life RL settings. As seen in 
Table~\ref{tab:main}, we find that \algo achieves the lowest average and median number of steps as well as highest number of successes on all four domains. In particular, \algo takes 20-40\% fewer steps on average as the next best performing method. 
GAIL-s and GAIL-sa also outperform finetuning SAC across three of the four domains. These results demonstrate the suitability of distribution-matching approaches over RL finetuning in the \slrl setting. 
We see that guidance particularly towards a good state distribution is important, as we compare to finetuning SAC with an exploration bonus through random network distillation (\cite{burda2018exploration}). From Table~\ref{tab:main}, although RND may improve performance, particularly in the Tabletop domain, it generally does not perform as well as the distribution matching approaches, especially \algo, showing that simply increasing exploration is not enough. Furthermore, these results show that the additional shaping provided by weighting the prior data by $Q$-value when training the discriminator can significantly improve guidance towards the goal. While GAIL gives equal weight to all transitions along an optimal path, the agent in \algo is consistently guided towards states in the prior data with higher $Q$-value, leading to more efficient and reliable single-life task completion. Additional comparisons are presented in the Appendix.



\begin{table}[t]
    \center
    \resizebox{\textwidth}{!}{%
    \renewcommand{\arraystretch}{1.2}
\begin{tabular}{lcccc|lcccc}
\toprule
                              & Method & Avg $\pm$ Std error & Success / 10 & Median & \multicolumn{1}{c}{}        & Method & Avg $\pm$ Std error & Success / 10 & Median \\ 
                              \midrule
\multicolumn{1}{c}{Tabletop}  & GAIL-s & 83.2k   $\pm$ 23.8k     & 8            & 75.6k  & \multicolumn{1}{c}{Cheetah} & GAIL-s & 99.2k   $\pm$ 23.0k     & 7            & 77.4k  \\
 & GAIL-sa     & 61.5k  $\pm$ 28.7k & 7 & \textbf{2.4k}   &  & GAIL-sa     & 102.0k $\pm$ 19.3k & 8 & 85.6k  \\
 & SAC     & 123.7k  $\pm$ 25.5k & 7 & 157.2k   &  & SAC     & 128.9k $\pm$ 23.9k & 5 & 138.2k  \\
 & SAC-RND     & 94.8k  $\pm$ 26.9k & 7 & 51.4k   &  & SAC-RND     & 158.4k $\pm$ 20.5k & 3 & 200.0k  \\
 & \algo (ours)   & \textbf{44.4k  $\pm$ 24.6k}  & \textbf{9} & 8.9k  &  & \algo (ours)   & \textbf{74.3k  $\pm$ 9.5k} & \textbf{10} & \textbf{68.9k}  \\
 \midrule
\multicolumn{1}{c}{Pointmass} & GAIL-s & 100.9k  $\pm$ 33.0k     & 5            & 101.9k & \multicolumn{1}{c}{Kitchen} & GAIL-s & 111.3k  $\pm$ 27.9k    & 6            & 122.1k \\
 & GAIL-sa     & 140.4k $\pm$ 30.3k & 3 & 200.0k &  & GAIL-sa     & 127.8k $\pm$ 26.9k & 5 & 189.1k \\
  & SAC     & 198.1k  $\pm$ 1.9k & 1 & 200.0k   &  & SAC     & 112.5k $\pm$ 25.8k & 6 & 122.3k  \\
 & SAC-RND     & 184.1k  $\pm$ 15.9k & 1 & 200.0k   &  & SAC-RND     & 132.7k $\pm$ 26.2k & 4 & 200.0k  \\
 & \algo (ours)   & \textbf{61.0k  $\pm$ 28.8k} & \textbf{8} & \textbf{1.7k}   &  & \algo (ours)   & \textbf{79.9k $\pm$ 21.7k} & \textbf{8} & \textbf{59.3k} \\
 \bottomrule
\end{tabular}
}
\vspace{0.3cm}
\caption{\small We evaluate the performance of \algo to finetuning SAC and GAIL in our four environments using mixed data collected through RL as prior data. We omit the results of Behavior Cloning (BC), as it is unsuccessful at completing the task in every domain due to the distribution shift. We find that GAIL outperforms finetuning SAC and SAC-RND in 3 out of 4 domains, and \algo outperforms both GAIL variants on all 4 domains on the average number of steps before task completion. All methods are evaluated over 10 seeds.}
\label{tab:main}
\vspace{-5mm}
\end{table}

\vspace{5mm}
\subsection{Recovering from novel situations in SLRL}

\begin{wrapfigure}{r}{0.6\linewidth}
    \centering
    \vspace{-0.5cm}
    \includegraphics[width=0.6\textwidth]{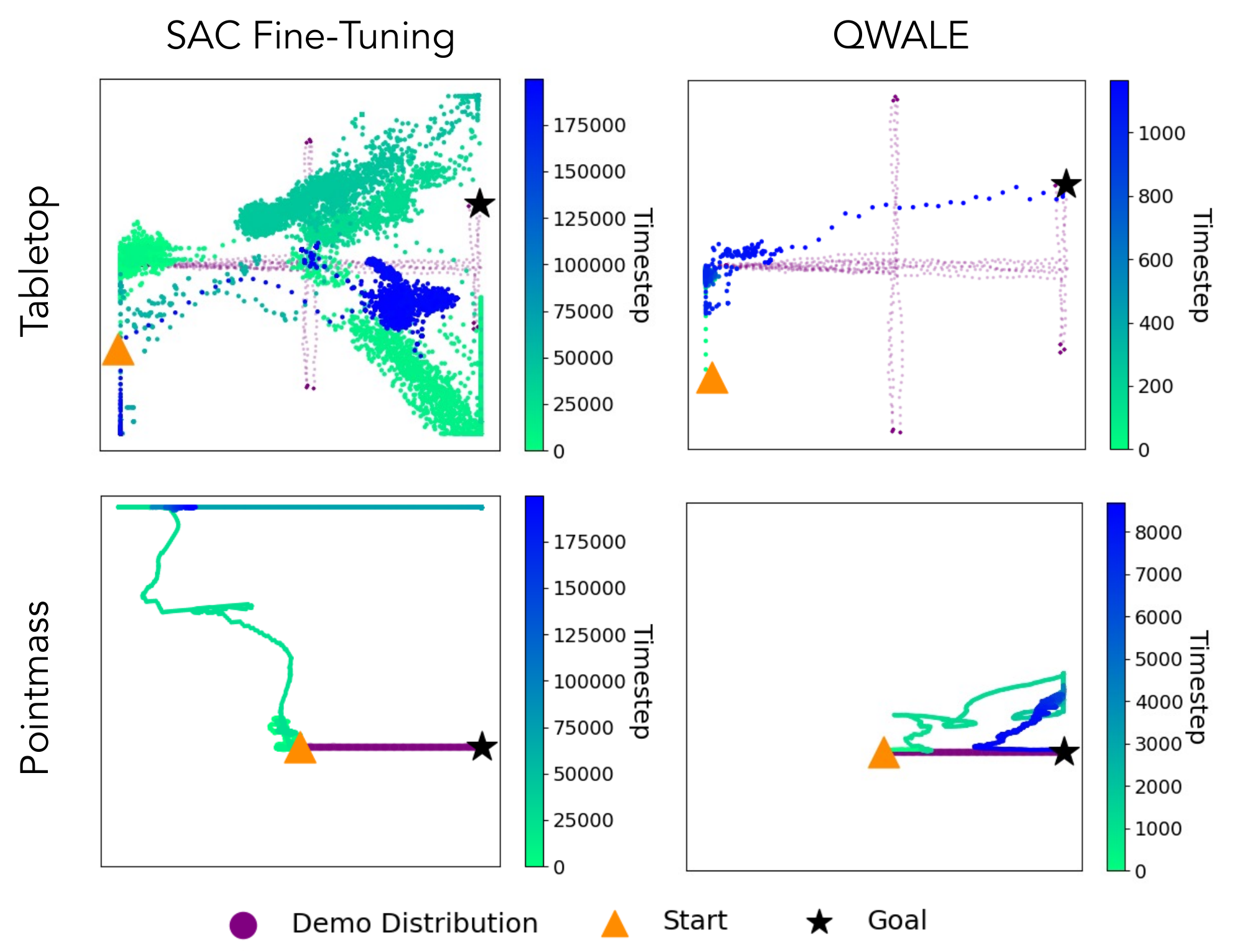}
    \caption{
        \small We visualize the online state visitation of the mug in the Tabletop (top) and the agent in the Pointmass (bottom) environments of a single-life trial using SAC finetuning (left) and \algo (right). The expert states are in purple, and we color the trajectories green to blue according to timestep for both methods. See Section~\ref{fig:visitation-reward} for plots colored according to reward. In both environments, the SAC fine-tuning agent fails to recover from novel states, but is guided towards task completion with \algo.
    }
    \label{fig:visitation}
\end{wrapfigure}
Next, to answer question (2), we analyze how \algo helps agents learn to recover from novel situations in \slrl. 
First, we analyze the state visitation of SAC finetuning in the online phase for the Tabletop and Pointmass domains.
A key challenge of \slrl (and also fully autonomous RL in general) is that if the agent falls off of
the distribution, it cannot rely on resets to get back on track. Since there is a gap between the source and target domains,
the agent will inevitably find itself in states that are out of distribution from the prior data.
A value function pre-trained on the source data could in principle be used to evaluate states and guide the agent back towards good states, but it will be inaccurate on states outside of the prior data \cite{fu2019diagnosing}; then, when the value of some of those out-of-distributions states is overestimated, the value function may misguide the policy away from good states.
Hence, fine-tuning a pre-trained value function via online RL
will not explicitly encourage the agent to get back on distribution, especially in sparse reward envs. As a result, the agent may spend a lot of time (perhaps infinite time) drifting once it falls out of distribution, which we see occurs in Figure~\ref{fig:visitation}.

On the other hand, distribution-matching methods like \algo will explicitly encourage the agent to get back on distribution, by giving higher rewards on distribution than off distribution. 
In Figure~\ref{fig:visitation}, we visualize \algo's state visitation in the Tabletop and Pointmass domains throughout a single lifetime. We color the trajectories according to timestep for both methods, and include the trajectories colored by reward (discriminator score) in Figure~\ref{fig:visitation-reward}. The coloring in the timestep-colored plots is highly correlated with that in the reward-colored plots, showing how the reward helps the agent recover from unfamiliar states and gradually guides the agent towards the goal. In particular, when the agent is out of distribution, the agent is incentivized to explore states that will lead it closer back to the prior state distribution, and when the agent is within distribution, it is incentivized to move to states closer to the goal, leading to efficient task completion.

\subsection{Using demos as prior data}
\label{sec:demo-prior}

Finally, we evaluate the performance of different discriminator-based approaches in the two \slrl problem settings--Tabletop and Pointmass--where demonstration data is available as prior data. We compare using AIL with a state-only discriminator (GAIL-s) as well as with a state-action discriminator (GAIL-sa) to our proposed $Q$-weighted discriminator method (\algo). With the latter method, we have access to the same $Q$-function pretrained when collecting prior data using standard RL for the mixed data experiments above, but we do not initialize any of the algorithms at test time with the pretrained policy and critic weights. 

\begin{figure*}[t!]
    \centering
    \includegraphics[width=\textwidth]{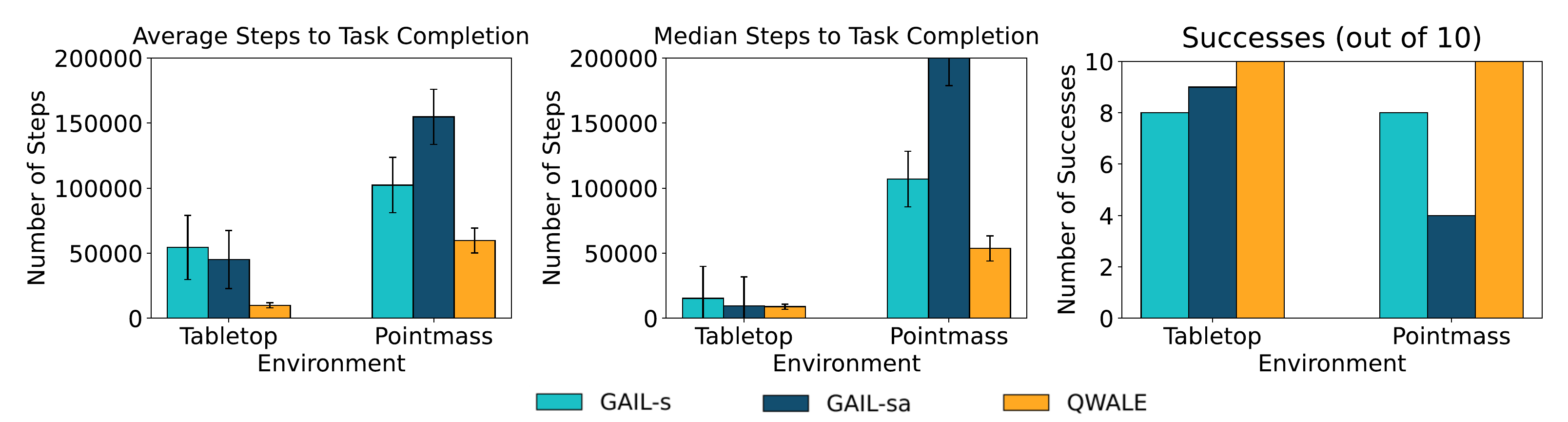}
    \caption{\small
        Discriminator-based approaches to single-life RL with expert demonstrations as the prior data. Given demonstration data, in both the Tabletop and Pointmass domains,  \algo matches or significantly outperforms both GAIL variants on all metrics.
    }
    \label{fig:demodist}
\end{figure*}

From Figure~\ref{fig:demodist}, the GAIL variants are both able to consistently solve the task in the Tabletop domain, but unsurprisingly, GAIL-sa does especially poorly in the Pointmass domain, where the dynamics have changed at test time. Compared to the two GAIL variants, \algo gives a significant improvement in both domains. These results demonstrate how more detailed reward shaping towards the completion of the desired task can be helpful in the \slrl setting even with demonstrations as prior data. Moreover, comparing these results with those in Table~\ref{tab:main}, while access to expert data as prior data unsurprisingly improves the performance of GAIL methods, it can also improve the performance of \algo.

\section{Conclusion}
\label{sec:conclusion}

In this paper, we formalized and studied the problem setting of single-life reinforcement learning: a setting where an agent needs to autonomously complete a task once in a single trial while drawing upon prior experience from a related environment. We found that standard fine-tuning via RL is ill-suited for this problem because the algorithm struggles to recover from mistakes and novel situations. We hypothesized that this happens since resets in episodic RL prevent algorithms from needing to recover, whereas single-life RL and continuing settings in general do demand the agent to find its way back to good states on its own. We then postulated that distribution matching methods that aim to match the distribution of related prior data may help agents recover via reward shaping, and presented a new distribution matching method, \algo, that weights examples by their $Q$-value. Our experiments verified that distribution matching approaches indeed do make better use of prior data, and that \algo outperforms prior distribution matching methods on four single-life RL problems.

While \algo can efficiently complete novel target tasks in a single episode without any interventions, important limitations remain. No algorithm, including \algo, was able to complete the target task with 100\% success, indicating that future works should aim to improve an algorithm's ability to solve tasks consistently. Moreover, the methods that we evaluated all used a pre-trained policy and value function from the source domain, which may be difficult to obtain in some source scenarios, as opposed to only obtaining some demonstrations or offline data. Finally, it would be interesting to explore problems with greater degrees of novelty between the source and target environments. In our experimental domains, the reward and state space between the source and target MDPs were assumed to stay the same. Eventually, we hope that this work may be extended to the case where the reward may be different in the online trial from the prior data.
We expect that such settings would place even greater importance on autonomy and exploration, requiring sophisticated strategies for adapting online to novelty. We hope that future work can explore these interesting questions and continue to make progress on allowing RL agents to autonomously complete tasks within a single lifetime.

\section*{Acknowledgments}
We thank members of the IRIS lab for helpful discussions on this project. This research was supported by Google and the Office of Naval Research (ONR) via grants N00014-21-1-2685 and N00014-20-1-2675. Annie Chen is supported by an NSF graduate student fellowship. Chelsea Finn is a CIFAR fellow.

\bibliographystyle{plainnat}
\bibliography{references}

\begin{thebibliography}{71}
\providecommand{\natexlab}[1]{#1}
\providecommand{\url}[1]{\texttt{#1}}
\expandafter\ifx\csname urlstyle\endcsname\relax
  \providecommand{\doi}[1]{doi: #1}\else
  \providecommand{\doi}{doi: \begingroup \urlstyle{rm}\Url}\fi

\bibitem[Abdolmaleki et~al.(2018)Abdolmaleki, Springenberg, Tassa, Munos,
  Heess, and Riedmiller]{abdolmaleki2018maximum}
Abbas Abdolmaleki, Jost~Tobias Springenberg, Yuval Tassa, Remi Munos, Nicolas
  Heess, and Martin Riedmiller.
\newblock Maximum a posteriori policy optimisation.
\newblock \emph{arXiv preprint arXiv:1806.06920}, 2018.

\bibitem[Argall et~al.(2009)Argall, Chernova, Veloso, and
  Browning]{argall2009survey}
Brenna~D Argall, Sonia Chernova, Manuela Veloso, and Brett Browning.
\newblock A survey of robot learning from demonstration.
\newblock \emph{Robotics and autonomous systems}, 57\penalty0 (5):\penalty0
  469--483, 2009.

\bibitem[Beliaev et~al.(2022)Beliaev, Shih, Ermon, Sadigh, and
  Pedarsani]{beliaev2022imitation}
Mark Beliaev, Andy Shih, Stefano Ermon, Dorsa Sadigh, and Ramtin Pedarsani.
\newblock Imitation learning by estimating expertise of demonstrators.
\newblock \emph{arXiv preprint arXiv:2202.01288}, 2022.

\bibitem[Brys et~al.(2015)Brys, Harutyunyan, Suay, Chernova, Taylor, and
  Now{\'e}]{brys2015reinforcement}
Tim Brys, Anna Harutyunyan, Halit~Bener Suay, Sonia Chernova, Matthew~E Taylor,
  and Ann Now{\'e}.
\newblock Reinforcement learning from demonstration through shaping.
\newblock In \emph{Twenty-fourth international joint conference on artificial
  intelligence}, 2015.

\bibitem[Burda et~al.(2018)Burda, Edwards, Storkey, and
  Klimov]{burda2018exploration}
Yuri Burda, Harrison Edwards, Amos Storkey, and Oleg Klimov.
\newblock Exploration by random network distillation.
\newblock \emph{arXiv preprint arXiv:1810.12894}, 2018.

\bibitem[Cao et~al.(2022)Cao, Wang, and Sadigh]{cao2022learning}
Zhangjie Cao, Zihan Wang, and Dorsa Sadigh.
\newblock Learning from imperfect demonstrations via adversarial confidence
  transfer.
\newblock \emph{arXiv preprint arXiv:2202.02967}, 2022.

\bibitem[Eysenbach et~al.(2017)Eysenbach, Gu, Ibarz, and
  Levine]{eysenbach2017leave}
Benjamin Eysenbach, Shixiang Gu, Julian Ibarz, and Sergey Levine.
\newblock Leave no trace: Learning to reset for safe and autonomous
  reinforcement learning.
\newblock \emph{arXiv preprint arXiv:1711.06782}, 2017.

\bibitem[Eysenbach et~al.(2020)Eysenbach, Asawa, Chaudhari, Levine, and
  Salakhutdinov]{eysenbach2020off}
Benjamin Eysenbach, Swapnil Asawa, Shreyas Chaudhari, Sergey Levine, and Ruslan
  Salakhutdinov.
\newblock Off-dynamics reinforcement learning: Training for transfer with
  domain classifiers.
\newblock \emph{arXiv preprint arXiv:2006.13916}, 2020.

\bibitem[Finn et~al.(2016)Finn, Levine, and Abbeel]{finn2016guided}
Chelsea Finn, Sergey Levine, and Pieter Abbeel.
\newblock Guided cost learning: Deep inverse optimal control via policy
  optimization.
\newblock In \emph{International conference on machine learning}, pages 49--58.
  PMLR, 2016.

\bibitem[Finn et~al.(2017)Finn, Abbeel, and Levine]{finn2017model}
Chelsea Finn, Pieter Abbeel, and Sergey Levine.
\newblock Model-agnostic meta-learning for fast adaptation of deep networks.
\newblock In \emph{International conference on machine learning}, pages
  1126--1135. PMLR, 2017.

\bibitem[Finn et~al.(2019)Finn, Rajeswaran, Kakade, and Levine]{finn2019online}
Chelsea Finn, Aravind Rajeswaran, Sham Kakade, and Sergey Levine.
\newblock Online meta-learning.
\newblock In \emph{International Conference on Machine Learning}, pages
  1920--1930. PMLR, 2019.

\bibitem[Fu et~al.(2017)Fu, Luo, and Levine]{fu2017learning}
Justin Fu, Katie Luo, and Sergey Levine.
\newblock Learning robust rewards with adversarial inverse reinforcement
  learning.
\newblock \emph{arXiv preprint arXiv:1710.11248}, 2017.

\bibitem[Fu et~al.(2019)Fu, Kumar, Soh, and Levine]{fu2019diagnosing}
Justin Fu, Aviral Kumar, Matthew Soh, and Sergey Levine.
\newblock Diagnosing bottlenecks in deep q-learning algorithms.
\newblock In \emph{International Conference on Machine Learning}, pages
  2021--2030. PMLR, 2019.

\bibitem[Ghasemipour et~al.(2020)Ghasemipour, Zemel, and
  Gu]{ghasemipour2020divergence}
Seyed Kamyar~Seyed Ghasemipour, Richard Zemel, and Shixiang Gu.
\newblock A divergence minimization perspective on imitation learning methods.
\newblock In \emph{Conference on Robot Learning}, pages 1259--1277. PMLR, 2020.

\bibitem[Gupta et~al.(2019)Gupta, Kumar, Lynch, Levine, and
  Hausman]{gupta2019relay}
Abhishek Gupta, Vikash Kumar, Corey Lynch, Sergey Levine, and Karol Hausman.
\newblock Relay policy learning: Solving long-horizon tasks via imitation and
  reinforcement learning.
\newblock \emph{arXiv preprint arXiv:1910.11956}, 2019.

\bibitem[Gupta et~al.(2021)Gupta, Yu, Zhao, Kumar, Rovinsky, Xu, Devlin, and
  Levine]{gupta2021reset}
Abhishek Gupta, Justin Yu, Tony~Z Zhao, Vikash Kumar, Aaron Rovinsky, Kelvin
  Xu, Thomas Devlin, and Sergey Levine.
\newblock Reset-free reinforcement learning via multi-task learning: Learning
  dexterous manipulation behaviors without human intervention.
\newblock In \emph{2021 IEEE International Conference on Robotics and
  Automation (ICRA)}, pages 6664--6671. IEEE, 2021.

\bibitem[Gupta et~al.(2022)Gupta, Lynch, Kinman, Peake, Levine, and
  Hausman]{gupta2022bootstrapped}
Abhishek Gupta, Corey Lynch, Brandon Kinman, Garrett Peake, Sergey Levine, and
  Karol Hausman.
\newblock Bootstrapped autonomous practicing via multi-task reinforcement
  learning.
\newblock \emph{arXiv preprint arXiv:2203.15755}, 2022.

\bibitem[Haarnoja et~al.(2018)Haarnoja, Zhou, Abbeel, and
  Levine]{haarnoja2018soft}
Tuomas Haarnoja, Aurick Zhou, Pieter Abbeel, and Sergey Levine.
\newblock Soft actor-critic: Off-policy maximum entropy deep reinforcement
  learning with a stochastic actor.
\newblock In \emph{International conference on machine learning}, pages
  1861--1870. PMLR, 2018.

\bibitem[Han et~al.(2015)Han, Levine, and Abbeel]{han2015learning}
Weiqiao Han, Sergey Levine, and Pieter Abbeel.
\newblock Learning compound multi-step controllers under unknown dynamics.
\newblock In \emph{2015 IEEE/RSJ International Conference on Intelligent Robots
  and Systems (IROS)}, pages 6435--6442. IEEE, 2015.

\bibitem[Hansen et~al.(2020)Hansen, Jangir, Sun, Aleny{\`a}, Abbeel, Efros,
  Pinto, and Wang]{hansen2020self}
Nicklas Hansen, Rishabh Jangir, Yu~Sun, Guillem Aleny{\`a}, Pieter Abbeel,
  Alexei~A Efros, Lerrel Pinto, and Xiaolong Wang.
\newblock Self-supervised policy adaptation during deployment.
\newblock \emph{arXiv preprint arXiv:2007.04309}, 2020.

\bibitem[Hester et~al.(2018)Hester, Vecerik, Pietquin, Lanctot, Schaul, Piot,
  Horgan, Quan, Sendonaris, Osband, et~al.]{hester2018deep}
Todd Hester, Matej Vecerik, Olivier Pietquin, Marc Lanctot, Tom Schaul, Bilal
  Piot, Dan Horgan, John Quan, Andrew Sendonaris, Ian Osband, et~al.
\newblock Deep q-learning from demonstrations.
\newblock In \emph{Proceedings of the AAAI Conference on Artificial
  Intelligence}, volume~32, 2018.

\bibitem[Ho and Ermon(2016)]{ho2016generative}
Jonathan Ho and Stefano Ermon.
\newblock Generative adversarial imitation learning.
\newblock \emph{Advances in neural information processing systems}, 29, 2016.

\bibitem[Khetarpal et~al.(2020)Khetarpal, Riemer, Rish, and
  Precup]{khetarpal2020towards}
Khimya Khetarpal, Matthew Riemer, Irina Rish, and Doina Precup.
\newblock Towards continual reinforcement learning: A review and perspectives.
\newblock \emph{arXiv preprint arXiv:2012.13490}, 2020.

\bibitem[Kidambi et~al.(2020)Kidambi, Rajeswaran, Netrapalli, and
  Joachims]{kidambi2020morel}
Rahul Kidambi, Aravind Rajeswaran, Praneeth Netrapalli, and Thorsten Joachims.
\newblock Morel: Model-based offline reinforcement learning.
\newblock \emph{Advances in neural information processing systems},
  33:\penalty0 21810--21823, 2020.

\bibitem[Kim et~al.(2022)Kim, hyeon Park, Cho, and Kim]{kim2022automating}
Jigang Kim, J~hyeon Park, Daesol Cho, and H~Jin Kim.
\newblock Automating reinforcement learning with example-based resets.
\newblock \emph{IEEE Robotics and Automation Letters}, 2022.

\bibitem[Kostrikov et~al.(2018)Kostrikov, Agrawal, Dwibedi, Levine, and
  Tompson]{kostrikov2018discriminator}
Ilya Kostrikov, Kumar~Krishna Agrawal, Debidatta Dwibedi, Sergey Levine, and
  Jonathan Tompson.
\newblock Discriminator-actor-critic: Addressing sample inefficiency and reward
  bias in adversarial imitation learning.
\newblock \emph{arXiv preprint arXiv:1809.02925}, 2018.

\bibitem[Kumar et~al.(2020)Kumar, Zhou, Tucker, and
  Levine]{kumar2020conservative}
Aviral Kumar, Aurick Zhou, George Tucker, and Sergey Levine.
\newblock Conservative q-learning for offline reinforcement learning.
\newblock \emph{Advances in Neural Information Processing Systems},
  33:\penalty0 1179--1191, 2020.

\bibitem[Levine et~al.(2020)Levine, Kumar, Tucker, and Fu]{levine2020offline}
Sergey Levine, Aviral Kumar, George Tucker, and Justin Fu.
\newblock Offline reinforcement learning: Tutorial, review, and perspectives on
  open problems.
\newblock \emph{arXiv preprint arXiv:2005.01643}, 2020.

\bibitem[Lomonaco et~al.(2020)Lomonaco, Desai, Culurciello, and
  Maltoni]{Lomonaco_2020_CVPR_Workshops}
Vincenzo Lomonaco, Karan Desai, Eugenio Culurciello, and Davide Maltoni.
\newblock Continual reinforcement learning in 3d non-stationary environments.
\newblock In \emph{Proceedings of the IEEE/CVF Conference on Computer Vision
  and Pattern Recognition (CVPR) Workshops}, June 2020.

\bibitem[Mahadevan(1996)]{mahadevan1996average}
Sridhar Mahadevan.
\newblock Average reward reinforcement learning: Foundations, algorithms, and
  empirical results.
\newblock \emph{Machine learning}, 22\penalty0 (1):\penalty0 159--195, 1996.

\bibitem[Mehta et~al.(2020)Mehta, Diaz, Golemo, Pal, and
  Paull]{mehta2020active}
Bhairav Mehta, Manfred Diaz, Florian Golemo, Christopher~J Pal, and Liam Paull.
\newblock Active domain randomization.
\newblock In \emph{Conference on Robot Learning}, pages 1162--1176. PMLR, 2020.

\bibitem[Nagabandi et~al.(2018)Nagabandi, Clavera, Liu, Fearing, Abbeel,
  Levine, and Finn]{nagabandi2018learning}
Anusha Nagabandi, Ignasi Clavera, Simin Liu, Ronald~S Fearing, Pieter Abbeel,
  Sergey Levine, and Chelsea Finn.
\newblock Learning to adapt in dynamic, real-world environments through
  meta-reinforcement learning.
\newblock \emph{arXiv preprint arXiv:1803.11347}, 2018.

\bibitem[Nair et~al.(2018)Nair, McGrew, Andrychowicz, Zaremba, and
  Abbeel]{nair2018overcoming}
Ashvin Nair, Bob McGrew, Marcin Andrychowicz, Wojciech Zaremba, and Pieter
  Abbeel.
\newblock Overcoming exploration in reinforcement learning with demonstrations.
\newblock In \emph{2018 IEEE international conference on robotics and
  automation (ICRA)}, pages 6292--6299. IEEE, 2018.

\bibitem[Ng et~al.(2000)Ng, Russell, et~al.]{ng2000algorithms}
Andrew~Y Ng, Stuart~J Russell, et~al.
\newblock Algorithms for inverse reinforcement learning.
\newblock In \emph{Icml}, volume~1, page~2, 2000.

\bibitem[Nichol et~al.(2018)Nichol, Achiam, and Schulman]{nichol2018first}
Alex Nichol, Joshua Achiam, and John Schulman.
\newblock On first-order meta-learning algorithms.
\newblock \emph{arXiv preprint arXiv:1803.02999}, 2018.

\bibitem[Peng et~al.(2018)Peng, Andrychowicz, Zaremba, and Abbeel]{peng2018sim}
Xue~Bin Peng, Marcin Andrychowicz, Wojciech Zaremba, and Pieter Abbeel.
\newblock Sim-to-real transfer of robotic control with dynamics randomization.
\newblock In \emph{2018 IEEE international conference on robotics and
  automation (ICRA)}, pages 3803--3810. IEEE, 2018.

\bibitem[Peters et~al.(2010)Peters, Mulling, and Altun]{peters2010relative}
Jan Peters, Katharina Mulling, and Yasemin Altun.
\newblock Relative entropy policy search.
\newblock In \emph{Twenty-Fourth AAAI Conference on Artificial Intelligence},
  2010.

\bibitem[Rajeswaran et~al.(2017)Rajeswaran, Kumar, Gupta, Vezzani, Schulman,
  Todorov, and Levine]{rajeswaran2017learning}
Aravind Rajeswaran, Vikash Kumar, Abhishek Gupta, Giulia Vezzani, John
  Schulman, Emanuel Todorov, and Sergey Levine.
\newblock Learning complex dexterous manipulation with deep reinforcement
  learning and demonstrations.
\newblock \emph{arXiv preprint arXiv:1709.10087}, 2017.

\bibitem[Rolnick et~al.(2019)Rolnick, Ahuja, Schwarz, Lillicrap, and
  Wayne]{NEURIPS2019_fa7cdfad}
David Rolnick, Arun Ahuja, Jonathan Schwarz, Timothy Lillicrap, and Gregory
  Wayne.
\newblock Experience replay for continual learning.
\newblock In H.~Wallach, H.~Larochelle, A.~Beygelzimer, F.~d\textquotesingle
  Alch\'{e}-Buc, E.~Fox, and R.~Garnett, editors, \emph{Advances in Neural
  Information Processing Systems}, volume~32. Curran Associates, Inc., 2019.
\newblock URL
  \url{https://proceedings.neurips.cc/paper/2019/file/fa7cdfad1a5aaf8370ebeda47a1ff1c3-Paper.pdf}.

\bibitem[Ross et~al.(2011)Ross, Gordon, and Bagnell]{ross2011reduction}
St{\'e}phane Ross, Geoffrey Gordon, and Drew Bagnell.
\newblock A reduction of imitation learning and structured prediction to
  no-regret online learning.
\newblock In \emph{Proceedings of the fourteenth international conference on
  artificial intelligence and statistics}, pages 627--635. JMLR Workshop and
  Conference Proceedings, 2011.

\bibitem[Rusu et~al.(2016)Rusu, Rabinowitz, Desjardins, Soyer, Kirkpatrick,
  Kavukcuoglu, Pascanu, and Hadsell]{rusu2016progressive}
Andrei~A Rusu, Neil~C Rabinowitz, Guillaume Desjardins, Hubert Soyer, James
  Kirkpatrick, Koray Kavukcuoglu, Razvan Pascanu, and Raia Hadsell.
\newblock Progressive neural networks.
\newblock \emph{arXiv preprint arXiv:1606.04671}, 2016.

\bibitem[Sadeghi and Levine(2016)]{sadeghi2016cad2rl}
Fereshteh Sadeghi and Sergey Levine.
\newblock Cad2rl: Real single-image flight without a single real image.
\newblock \emph{arXiv preprint arXiv:1611.04201}, 2016.

\bibitem[Schwartz(1993)]{schwartz1993reinforcement}
Anton Schwartz.
\newblock A reinforcement learning method for maximizing undiscounted rewards.
\newblock In \emph{Proceedings of the tenth international conference on machine
  learning}, volume 298, pages 298--305, 1993.

\bibitem[Sharma et~al.(2021{\natexlab{a}})Sharma, Gupta, Levine, Hausman, and
  Finn]{sharma2021subgoal}
Archit Sharma, Abhishek Gupta, Sergey Levine, Karol Hausman, and Chelsea Finn.
\newblock Autonomous reinforcement learning via subgoal curricula.
\newblock In M.~Ranzato, A.~Beygelzimer, Y.~Dauphin, P.S. Liang, and J.~Wortman
  Vaughan, editors, \emph{Advances in Neural Information Processing Systems},
  volume~34, pages 18474--18486. Curran Associates, Inc., 2021{\natexlab{a}}.
\newblock URL
  \url{https://proceedings.neurips.cc/paper/2021/file/99c83c904d0d64fbef50d919a5c66a80-Paper.pdf}.

\bibitem[Sharma et~al.(2021{\natexlab{b}})Sharma, Xu, Sardana, Gupta, Hausman,
  Levine, and Finn]{sharma2021autonomous}
Archit Sharma, Kelvin Xu, Nikhil Sardana, Abhishek Gupta, Karol Hausman, Sergey
  Levine, and Chelsea Finn.
\newblock Autonomous reinforcement learning: Formalism and benchmarking.
\newblock \emph{arXiv preprint arXiv:2112.09605}, 2021{\natexlab{b}}.

\bibitem[Sharma et~al.(2022)Sharma, Ahmad, and Finn]{sharma2022state}
Archit Sharma, Rehaan Ahmad, and Chelsea Finn.
\newblock A state-distribution matching approach to non-episodic reinforcement
  learning.
\newblock \emph{arXiv preprint arXiv:2205.05212}, 2022.

\bibitem[Singh et~al.(2019)Singh, Yang, Hartikainen, Finn, and
  Levine]{singh2019end}
Avi Singh, Larry Yang, Kristian Hartikainen, Chelsea Finn, and Sergey Levine.
\newblock End-to-end robotic reinforcement learning without reward engineering.
\newblock \emph{arXiv preprint arXiv:1904.07854}, 2019.

\bibitem[Sun and Ma(2019)]{sun2019adversarial}
Mingfei Sun and Xiaojuan Ma.
\newblock Adversarial imitation learning from incomplete demonstrations.
\newblock \emph{arXiv preprint arXiv:1905.12310}, 2019.

\bibitem[Sutton and Barto(2018)]{sutton2018reinforcement}
Richard~S Sutton and Andrew~G Barto.
\newblock \emph{Reinforcement learning: An introduction}.
\newblock MIT press, 2018.

\bibitem[Syed et~al.(2008)Syed, Bowling, and Schapire]{syed2008apprenticeship}
Umar Syed, Michael Bowling, and Robert~E Schapire.
\newblock Apprenticeship learning using linear programming.
\newblock In \emph{Proceedings of the 25th international conference on Machine
  learning}, pages 1032--1039, 2008.

\bibitem[Tobin et~al.(2017)Tobin, Fong, Ray, Schneider, Zaremba, and
  Abbeel]{tobin2017domain}
Josh Tobin, Rachel Fong, Alex Ray, Jonas Schneider, Wojciech Zaremba, and
  Pieter Abbeel.
\newblock Domain randomization for transferring deep neural networks from
  simulation to the real world.
\newblock In \emph{2017 IEEE/RSJ international conference on intelligent robots
  and systems (IROS)}, pages 23--30. IEEE, 2017.

\bibitem[Torabi et~al.(2019)Torabi, Warnell, and Stone]{torabi2019adversarial}
Faraz Torabi, Garrett Warnell, and Peter Stone.
\newblock Adversarial imitation learning from state-only demonstrations.
\newblock In \emph{Proceedings of the 18th International Conference on
  Autonomous Agents and MultiAgent Systems}, pages 2229--2231, 2019.

\bibitem[Vecerik et~al.(2017)Vecerik, Hester, Scholz, Wang, Pietquin, Piot,
  Heess, Roth{\"o}rl, Lampe, and Riedmiller]{vecerik2017leveraging}
Mel Vecerik, Todd Hester, Jonathan Scholz, Fumin Wang, Olivier Pietquin, Bilal
  Piot, Nicolas Heess, Thomas Roth{\"o}rl, Thomas Lampe, and Martin Riedmiller.
\newblock Leveraging demonstrations for deep reinforcement learning on robotics
  problems with sparse rewards.
\newblock \emph{arXiv preprint arXiv:1707.08817}, 2017.

\bibitem[Wang et~al.(2018)Wang, Xiong, Han, Liu, Zhang,
  et~al.]{wang2018exponentially}
Qing Wang, Jiechao Xiong, Lei Han, Han Liu, Tong Zhang, et~al.
\newblock Exponentially weighted imitation learning for batched historical
  data.
\newblock \emph{Advances in Neural Information Processing Systems}, 31, 2018.

\bibitem[Wang et~al.(2020{\natexlab{a}})Wang, Ciliberto, Amadori, and
  Demiris]{wang2020support}
Ruohan Wang, Carlo Ciliberto, Pierluigi Amadori, and Yiannis Demiris.
\newblock Support-weighted adversarial imitation learning.
\newblock \emph{arXiv preprint arXiv:2002.08803}, 2020{\natexlab{a}}.

\bibitem[Wang et~al.(2021{\natexlab{a}})Wang, Xu, and Du]{wang2021robust}
Yunke Wang, Chang Xu, and Bo~Du.
\newblock Robust adversarial imitation learning via adaptively-selected
  demonstrations.
\newblock In Zhi-Hua Zhou, editor, \emph{Proceedings of the Thirtieth
  International Joint Conference on Artificial Intelligence, {IJCAI-21}}, pages
  3155--3161. International Joint Conferences on Artificial Intelligence
  Organization, 2021{\natexlab{a}}.

\bibitem[Wang et~al.(2021{\natexlab{b}})Wang, Xu, Du, and
  Lee]{wang2021learning}
Yunke Wang, Chang Xu, Bo~Du, and Honglak Lee.
\newblock Learning to weight imperfect demonstrations.
\newblock In \emph{International Conference on Machine Learning}, pages
  10961--10970. PMLR, 2021{\natexlab{b}}.

\bibitem[Wang et~al.(2020{\natexlab{b}})Wang, Novikov, Zolna, Merel,
  Springenberg, Reed, Shahriari, Siegel, Gulcehre, Heess,
  et~al.]{wang2020critic}
Ziyu Wang, Alexander Novikov, Konrad Zolna, Josh~S Merel, Jost~Tobias
  Springenberg, Scott~E Reed, Bobak Shahriari, Noah Siegel, Caglar Gulcehre,
  Nicolas Heess, et~al.
\newblock Critic regularized regression.
\newblock \emph{Advances in Neural Information Processing Systems},
  33:\penalty0 7768--7778, 2020{\natexlab{b}}.

\bibitem[Wei et~al.(2020)Wei, Jahromi, Luo, Sharma, and Jain]{pmlr-v119-wei20c}
Chen-Yu Wei, Mehdi~Jafarnia Jahromi, Haipeng Luo, Hiteshi Sharma, and Rahul
  Jain.
\newblock Model-free reinforcement learning in infinite-horizon average-reward
  {M}arkov decision processes.
\newblock In Hal~Daumé III and Aarti Singh, editors, \emph{Proceedings of the
  37th International Conference on Machine Learning}, volume 119 of
  \emph{Proceedings of Machine Learning Research}, pages 10170--10180. PMLR,
  13--18 Jul 2020.

\bibitem[Wu et~al.(2019{\natexlab{a}})Wu, Tucker, and Nachum]{wu2019behavior}
Yifan Wu, George Tucker, and Ofir Nachum.
\newblock Behavior regularized offline reinforcement learning.
\newblock \emph{arXiv preprint arXiv:1911.11361}, 2019{\natexlab{a}}.

\bibitem[Wu et~al.(2019{\natexlab{b}})Wu, Charoenphakdee, Bao, Tangkaratt, and
  Sugiyama]{wu2019imitation}
Yueh-Hua Wu, Nontawat Charoenphakdee, Han Bao, Voot Tangkaratt, and Masashi
  Sugiyama.
\newblock Imitation learning from imperfect demonstration.
\newblock In \emph{International Conference on Machine Learning}, pages
  6818--6827. PMLR, 2019{\natexlab{b}}.

\bibitem[Xie and Finn(2021)]{xie2021lifelong}
Annie Xie and Chelsea Finn.
\newblock Lifelong robotic reinforcement learning by retaining experiences.
\newblock \emph{arXiv preprint arXiv:2109.09180}, 2021.

\bibitem[Xie et~al.(2020)Xie, Harrison, and Finn]{xie2020deep}
Annie Xie, James Harrison, and Chelsea Finn.
\newblock Deep reinforcement learning amidst lifelong non-stationarity.
\newblock \emph{arXiv preprint arXiv:2006.10701}, 2020.

\bibitem[Xu and Zhu(2018)]{xu2018reinforced}
Ju~Xu and Zhanxing Zhu.
\newblock Reinforced continual learning.
\newblock In S.~Bengio, H.~Wallach, H.~Larochelle, K.~Grauman, N.~Cesa-Bianchi,
  and R.~Garnett, editors, \emph{Advances in Neural Information Processing
  Systems}, volume~31. Curran Associates, Inc., 2018.
\newblock URL
  \url{https://proceedings.neurips.cc/paper/2018/file/cee631121c2ec9232f3a2f028ad5c89b-Paper.pdf}.

\bibitem[Yoneda et~al.(2021)Yoneda, Yang, Walter, and
  Stadie]{yoneda2021invariance}
Takuma Yoneda, Ge~Yang, Matthew~R Walter, and Bradly Stadie.
\newblock Invariance through inference.
\newblock \emph{arXiv preprint arXiv:2112.08526}, 2021.

\bibitem[Zhang et~al.(2017)Zhang, Cisse, Dauphin, and
  Lopez-Paz]{zhang2017mixup}
Hongyi Zhang, Moustapha Cisse, Yann~N Dauphin, and David Lopez-Paz.
\newblock mixup: Beyond empirical risk minimization.
\newblock \emph{arXiv preprint arXiv:1710.09412}, 2017.

\bibitem[Zhu et~al.(2020{\natexlab{a}})Zhu, Yu, Gupta, Shah, Hartikainen,
  Singh, Kumar, and Levine]{zhu2020ingredients}
Henry Zhu, Justin Yu, Abhishek Gupta, Dhruv Shah, Kristian Hartikainen, Avi
  Singh, Vikash Kumar, and Sergey Levine.
\newblock The ingredients of real-world robotic reinforcement learning.
\newblock \emph{arXiv preprint arXiv:2004.12570}, 2020{\natexlab{a}}.

\bibitem[Zhu et~al.(2020{\natexlab{b}})Zhu, Lin, Dai, and Zhou]{zhu2020off}
Zhuangdi Zhu, Kaixiang Lin, Bo~Dai, and Jiayu Zhou.
\newblock Off-policy imitation learning from observations.
\newblock In \emph{the Thirty-fourth Annual Conference on Neural Information
  Processing Systems (NeurIPS 2020)}, 2020{\natexlab{b}}.

\bibitem[Ziebart et~al.(2008)Ziebart, Maas, Bagnell, Dey,
  et~al.]{ziebart2008maximum}
Brian~D Ziebart, Andrew~L Maas, J~Andrew Bagnell, Anind~K Dey, et~al.
\newblock Maximum entropy inverse reinforcement learning.
\newblock In \emph{AAAI}, volume~8, pages 1433--1438. Chicago, IL, USA, 2008.

\bibitem[Ziebart et~al.(2010)Ziebart, Bagnell, and Dey]{ziebart2010modeling}
Brian~D Ziebart, J~Andrew Bagnell, and Anind~K Dey.
\newblock Modeling interaction via the principle of maximum causal entropy.
\newblock In \emph{ICML}, 2010.

\bibitem[Zintgraf et~al.(2019)Zintgraf, Shiarlis, Igl, Schulze, Gal, Hofmann,
  and Whiteson]{zintgraf2019varibad}
Luisa Zintgraf, Kyriacos Shiarlis, Maximilian Igl, Sebastian Schulze, Yarin
  Gal, Katja Hofmann, and Shimon Whiteson.
\newblock Varibad: A very good method for bayes-adaptive deep rl via
  meta-learning.
\newblock \emph{arXiv preprint arXiv:1910.08348}, 2019.

\end{thebibliography}

\section*{Checklist}

\begin{enumerate}

\item For all authors...
\begin{enumerate}
  \item Do the main claims made in the abstract and introduction accurately reflect the paper's contributions and scope?
    \answerYes{}
  \item Did you describe the limitations of your work?
    \answerYes{See Section~\ref{sec:conclusion}.}
  \item Did you discuss any potential negative societal impacts of your work?
    \answerNA{We do not expect any negative societal impacts that are unique to this work compared to other works on reinforcement learning.}
  \item Have you read the ethics review guidelines and ensured that your paper conforms to them?
    \answerYes{}
\end{enumerate}

\item If you are including theoretical results...
\begin{enumerate}
  \item Did you state the full set of assumptions of all theoretical results?
    \answerNA{}
  \item Did you include complete proofs of all theoretical results?
    \answerNA{}
\end{enumerate}

\item If you ran experiments...
\begin{enumerate}
  \item Did you include the code, data, and instructions needed to reproduce the main experimental results (either in the supplemental material or as a URL)?
    \answerYes{Included in the supplemental material.}
  \item Did you specify all the training details (e.g., data splits, hyperparameters, how they were chosen)?
    \answerYes{Included in the supplemental material.}
        \item Did you report error bars (e.g., with respect to the random seed after running experiments multiple times)?
    \answerYes{See experiments in Section~\ref{sec:experiments}.}
        \item Did you include the total amount of compute and the type of resources used (e.g., type of GPUs, internal cluster, or cloud provider)?
    \answerYes{Included in the supplemental material.}
\end{enumerate}

\item If you are using existing assets (e.g., code, data, models) or curating/releasing new assets...
\begin{enumerate}
  \item If your work uses existing assets, did you cite the creators?
    \answerNA{}
  \item Did you mention the license of the assets?
    \answerNA{}
  \item Did you include any new assets either in the supplemental material or as a URL?
    \answerNA{}
  \item Did you discuss whether and how consent was obtained from people whose data you're using/curating?
    \answerNA{}
  \item Did you discuss whether the data you are using/curating contains personally identifiable information or offensive content?
    \answerNA{}
\end{enumerate}

\item If you used crowdsourcing or conducted research with human subjects...
\begin{enumerate}
  \item Did you include the full text of instructions given to participants and screenshots, if applicable?
    \answerNA{}
  \item Did you describe any potential participant risks, with links to Institutional Review Board (IRB) approvals, if applicable?
    \answerNA{}
  \item Did you include the estimated hourly wage paid to participants and the total amount spent on participant compensation?
    \answerNA{}
\end{enumerate}

\end{enumerate}

\newpage
\appendix
\section{Appendix}
\label{sec:appendix}

\subsection{Additional Method Justification}
\label{sec:app-method}

The key idea of \algo is to lead the agent to nearby states within distribution of the prior data if it is out of distribution and to nearby states closer to task completion if in distribution.
GAIL finds a policy whose occupancy measure minimizes the Jensen-Shannon divergence $\mathcal{D}_{\textrm{JS}}$ to the prior data. Since our objective in \slrl is to finish the task as soon as possible, and we may not be given expert demonstrations as prior data, we want to match the state-action pairs to those that lead to task completion. 

To this end, we consider the following optimization problem:
\begin{gather}
    \max_{\rho} \mathbb{E}_{\rho} \left[r(s, a)\right] = \max_{\rho} \int_{\mathcal{S} \times \mathcal{A}}  \rho(s, a) r(s, a) \dd s \dd a \label{eq:rl}\\
    \textrm{s.t.}\quad \int_{\mathcal{S} \times \mathcal{A}} \rho(s, a) \dd s \dd a = 1, \quad   \rho(s, a) \geq 0 \quad \forall (s, a) \in \mathcal{S} \times \mathcal{A}\label{eq:prob_valid}, \\
    \int_{\mathcal{A}} \rho(s, a) \dd a = (1-\gamma) \rho_0 (s) + \gamma \int_{\mathcal{S} \times \mathcal{A}} \rho(s', a') \mathcal{P}(s \mid s', a') \dd s' \dd a' \quad \forall s\in \mathcal{S},\label{eq:dyn_valid} \\
    \textrm{KL}\left(\rho(s, a) \mid\mid \rho_\beta(s, a)\right) \leq \epsilon. \label{eq:kl_constraint}
\end{gather}

Equations (~\ref{eq:rl}-~\ref{eq:dyn_valid}) represent the reinforcement learning problem stated in the formalism of convex optimization, where $\rho$ represents the state-action distribution being optimized, Eq~\ref{eq:prob_valid} constrains $\rho$ to be a valid probability distribution and Eq~\ref{eq:dyn_valid} constrains $\rho$ to be consistent with the MDP. The state-distribution $\rho^*$ of the expert policy $\pi_E$ is a solution of this problem. By introducing the constraint in Eq~\ref{eq:kl_constraint}, we want to encourage the optimal distribution $\rho^*_{\textrm{target}}(s, a)$ to be \textit{soft} by being close to the prior state-action distribution $\rho_\beta$ (the state-action distribution of $\mathcal{D}_{\textrm{prior}}$ corresponding to some behavior policy $\beta$). This problem has been studied in stochastic optimal control, particularly REPS~\citep{peters2010relative}. REPS shows that the solution to the optimization problem in Eq(~\ref{eq:rl}-~\ref{eq:kl_constraint}) has the form $\rho^*_{\textrm{target}}(s, a) \propto \rho_\beta(s, a) \exp (Q^{\pi_{\textrm{target}}}(s, a) - V^{\pi_{\textrm{target}}}(s))$, where the corresponding policy can be derived by using $\pi_{\textrm{target}} = \rho^*_{\textrm{target}}(s, a)/ \int_\mathcal{A}\rho^*_{\textrm{target}}(s, a') \dd a'$ from~\citep{syed2008apprenticeship}.

\subsection{Implementation Details and Hyperparameters}
In our experiments, we use soft actor-critic \citep{haarnoja2018soft} as our base RL algorithm. We use default hyperparameter values: a learning rate of 3e-4 for all networks, optimized using Adam, with a batch size of 256 sampled from the entire replay buffer (both prior and online data), a discount factor of 0.99. The policy and critic networks are MLPs with 2 fully-connected hidden layers of size 256. For all methods training a discriminator, it is parameterized as an MLP with 1 fully-connected hidden layer of size 128 and trained with a batch size of 512. During the online trial, 1000 steps are taken as initial collection steps before network updates begin. 
For all methods training a discriminator, we use mixup regularization \cite{zhang2017mixup} to reduce the brittleness of the discriminator. 

Following \citep{sharma2021autonomous}, we use a biased TD update, where $Q(s_t, a_t) \leftarrow r(s_t, a_t) + \gamma Q(s_{t+1}, a_{t+1})$ if $t$ is not a multiple of 100, and $Q(s_t, a_t) \leftarrow r(s_t, a_t)$ if it is. We use this update for all our evaluated methods online in order to improve stability. Since the online trial of single-life RL may have a large training horizon with hundreds of thousands of steps, this may lead to unstable bootstrapping, as for each $t$, $Q(s_t, a_t)$ bootstraps on $Q(s_{t+1}, a_{t+1}).$ Following \citep{sharma2022state}, for the auxiliary reward given by the discriminator $D$, we use $r(s, a) = -\log(1 - D(s))$ instead of $r(s, a) = \log D(s)$ to further improve stability. 

For \algo, the weighting of states is offset by a value $b$. This value may be treated like a constant hyperparameter and tuned. Adding this value changes the bias on the discriminator, which in effect adds a constant to the reward, though that constant changes over the course of training. In practice, to avoid having to tune $b$, we just use the value of the most recent state as $b$, i.e. $b = Q(s_t, a_t)$. To better interpret this value, with this weighting, $b$ is a baseline value capturing some notion of current progress.  Prior data tends to get small weights if they have worse value than the current state, so the agent is consistently incentivized to move towards states with higher value than its current state. 

For all experiments using prior data collected through RL, the agent was initialized at test time with the pretrained policy and critic. For \algo, a copy of that critic was frozen and used when calculating the weights for discriminator training. For all of the experiments with demonstration data in Section~\ref{sec:demo-prior}, the policy and critic were not initialized with any pretrained weights. 

\subsection{Environment \& Evaluation Details}
\textit{Tabletop-Organization.} The details for this environment are in \citep{sharma2021autonomous}. The state space consists of the gripper's $(x, y)$ position, the mug's $(x, y)$ position, the gripper's state (whether attached to the mug or not), and the current goal, for a total of 12 dimensions. The action space is 3 dimensional, consisting of a delta in the gripper's $(x, y)$ position as well as an automatic gripper that will attach to the mug if the gripper is close enough. The tabletop extends from -2.8 to 2.8 in both the $x$ and $y$ directions. In the prior data, which consists either of 10 demonstrations or 50000 transitions collected through RL after 350000 steps of training, the initial state always places the mug at position (2.5, 0.0), and the goal is to place the mug at one of the following locations: (-2.5, -1.0), (-2.5, 1.0), (0, 2.0), (0, -2). For the online trial when evaluating \slrl, the mug is placed either at (2.7, 1.5) or (2.7, -1.5) with additional uniform randomness between (-0.15, 0.15) in both directions. This environment is also goal-conditioned at test time and the goal is randomly set to be either (-2.5, -1.0) or (-2.5, 1.0). The reward is 1 when the mug is within 0.15 distance of its goal position (at which point the single life ends) and 0 everywhere else. 

\textit{Pointmass.} The Pointmass environment has a 6-dimensional state space consisting of the agent's $(x, y)$ position, its $(x, y)$ velocity, and the $(x, y)$ coordinates of the goal. The environment extends between -100 and 100 along the $x$ axis and between -200 and 200 along the $y$ axis. The action space is 2-D, consisting of the delta in both directions, clipped between -1 and 1 for a single action. The prior data consists of 3 demonstrations or 50000 transitions collected through RL after 350000 steps of training. The agent starts at (0, 0) and the goal is at (100, 0) for the prior data and online trial. During the online trial, a strong ``wind'' is introduced, where a random amount between 0.8 and 0.9 is added to the agent's $y$ coordinate and 0.2 is subtracted from the agent's $x$ coordinate at each step. The reward is 1 when the agent is within a distance of 2 of the goal position and 0 everywhere else. 

\textit{HalfCheetah.} The HalfCheetah environment has a state space with 18 dimensions, consisting of the position and velocity of each joint. The prior data consists of 50000 transitions collected through RL after 150000 steps of training. The reward is $r_t = \Delta x_t - 0.1 * ||a_t||^2_2$. At test time, 10 hurdles are included in the environment, spread between the x-coordinate of 7 and 260. The cheetah starts at 0 and its single life is considered successful when it gets to the coordinate 300, although the information about the hurdles or goal are not included in the state space.

\textit{Franka-Kitchen.} The Franka-Kitchen is adapted from \citep{gupta2019relay, sharma2021autonomous}. The state space consists of a 9 DoF position-controlled Franka-robot with a microwave and hinged cabinet. The prior data consists of 50000 transitions collected through standard episodic RL after 950000 steps of training, where one of the microwave or cabinet is open, and the task is to close that object. At test time, both are open, and the task is to close both objects. The reward function is equal to the sum of the Euclidean distance between the objects and their goal positions and the distance between the arm and its goal position.

\subsection{Additional Experiments}

We run three additional variants of SAC finetuning on all four experimental domains using mixed prior data. More specifically, we compare to (a) SAC-no online, in which we pretrain SAC in the source environment and evaluate in the target domain without additional finetuning, (b) SAC-scratch, where we train SAC completely from scratch in the target domain during its single episode, and (c) SAC-BC, where we fine-tune SAC like the comparison in Section~\ref{sec:rebuttal-exps} but add an additional behavior cloning loss on the offline data. 
\algo performs significantly better than three additional variants across all four domains. We find that SAC-no online performs poorly on all four test domains, which is unsurprising due to the distribution shift between the source and target environments. Adding a behavioral cloning loss does help in some domains but only if the dynamics are not changed--the method struggles in the Pointmass environment due to the changing dynamics at test time. SAC-scratch struggles compared to \algo and distribution-matching approaches but sometimes performs better than finetuning SAC, such as in the Tabletop domain, showing that the distribution shift is so significant in some domains that pretraining in the source domain is counterproductive when deployed in the target domain.

\begin{table}[t]
    \center
    \resizebox{\textwidth}{!}{%
    \renewcommand{\arraystretch}{1.2}
\begin{tabular}{lcccc|lcccc}
\toprule
                              & Method & Avg $\pm$ Std error & Success / 10 & Median & \multicolumn{1}{c}{}        & Method & Avg $\pm$ Std error & Success / 10 & Median \\ 
                              \midrule
\multicolumn{1}{c}{Tabletop}  & SAC-no online & 181.5k   $\pm$ 17.9k     & 2            & 200.0k  & \multicolumn{1}{c}{Cheetah} & SAC-no online & 143.7k   $\pm$ 28.6k     & 3            & 200.0k  \\
 & SAC-scratch     & 99.5k  $\pm$ 22.5k & 8 & 98.8k   &  & SAC-scratch    & 200.0k $\pm$ 0 & 0 & 200.0k  \\
 & SAC-BC     & 80.8k  $\pm$ 32.5k & 6 & 1.5k   &  & SAC-BC   & 93.5k $\pm$ 23.6k & 7 & 60.6k  \\
 \midrule
\multicolumn{1}{c}{Pointmass} & SAC-no online & 200.0k  $\pm$ 0     & 0            & 200.0k & \multicolumn{1}{c}{Kitchen} & SAC-no online & 200.0k  $\pm$ 0    & 0            & 200.0k \\
 & SAC-scratch     & 120.8  $\pm$ 18.5k & 8 & 112.9k   &  & SAC-scratch    & 155.0k $\pm$ 24.5k & 3 & 200.0k  \\
 & SAC-BC     & 200.0k  $\pm$ 0 & 0 & 200.0k   &  & SAC-BC   & 101.1k $\pm$ 28.1k & 6 & 71.7k  \\
 \bottomrule
\end{tabular}
}
\vspace{0.3cm}
\caption{\small We evaluate the performance of three additional SAC variants on our four experimental domains using mixed data collected through RL as prior data. \algo significantly outperforms all variants on all 4 domains on the average number of steps before task completion. We find that SAC without online learning performs particularly poorly and finetuning SAC with a behavioral cloning loss only helps if the dynamics are not changed. All methods are evaluated over 10 seeds.}
\label{tab:baselines}
\vspace{-5mm}
\end{table}

In addition, we run a comparison with a state-of-the-art reset-free algorithm, MEDAL~\cite{sharma2022state}, as well as an additional comparison with Progressive Networks~\cite{rusu2016progressive}, a strong continual learning method, in the SLRL setting in the Tabletop domain. We include these comparisons to demonstrate how the SLRL setting differs from typical autonomous RL and continual learning settings. Both methods perform poorly in SLRL setting, which is unsurprising because SLRL has the goal of completing the desired task a single time as quickly as possible, which is different from the other two settings. Thus, typical reset-free and continual learning algorithms may not be effective in the single-life setting.

\begin{table}[t]
    \center
    \resizebox{0.6\textwidth}{!}{%
    \renewcommand{\arraystretch}{1.2}
\begin{tabular}{lcccc}
\toprule
                              & Method & Avg $\pm$ Std error & Success / 10 & Median \\ 
                              \midrule
\multicolumn{1}{c}{Tabletop}  & MEDAL & 176.2k   $\pm$ 15.9k     & 3            & 200.0k  \\
 & Progressive Networks     & 159.0k  $\pm$ 21.2k & 3 & 200.0k   \\
 \bottomrule
\end{tabular}
}
\vspace{0.3cm}
\caption{\small We evaluate the performance of a state-of-the-art reset-free algorithm, MEDAL~\cite{sharma2022state}, and a continual learning approach, Progressive Networks~\cite{rusu2016progressive}, in the SLRL setting in the Tabletop domain. We find that both methods are not geared towards completing the desired task as quickly as possible.}
\label{tab:baselines}
\vspace{-5mm}
\end{table} 

\begin{figure*}[ht!]
    \centering
    \includegraphics[width=0.7\textwidth]{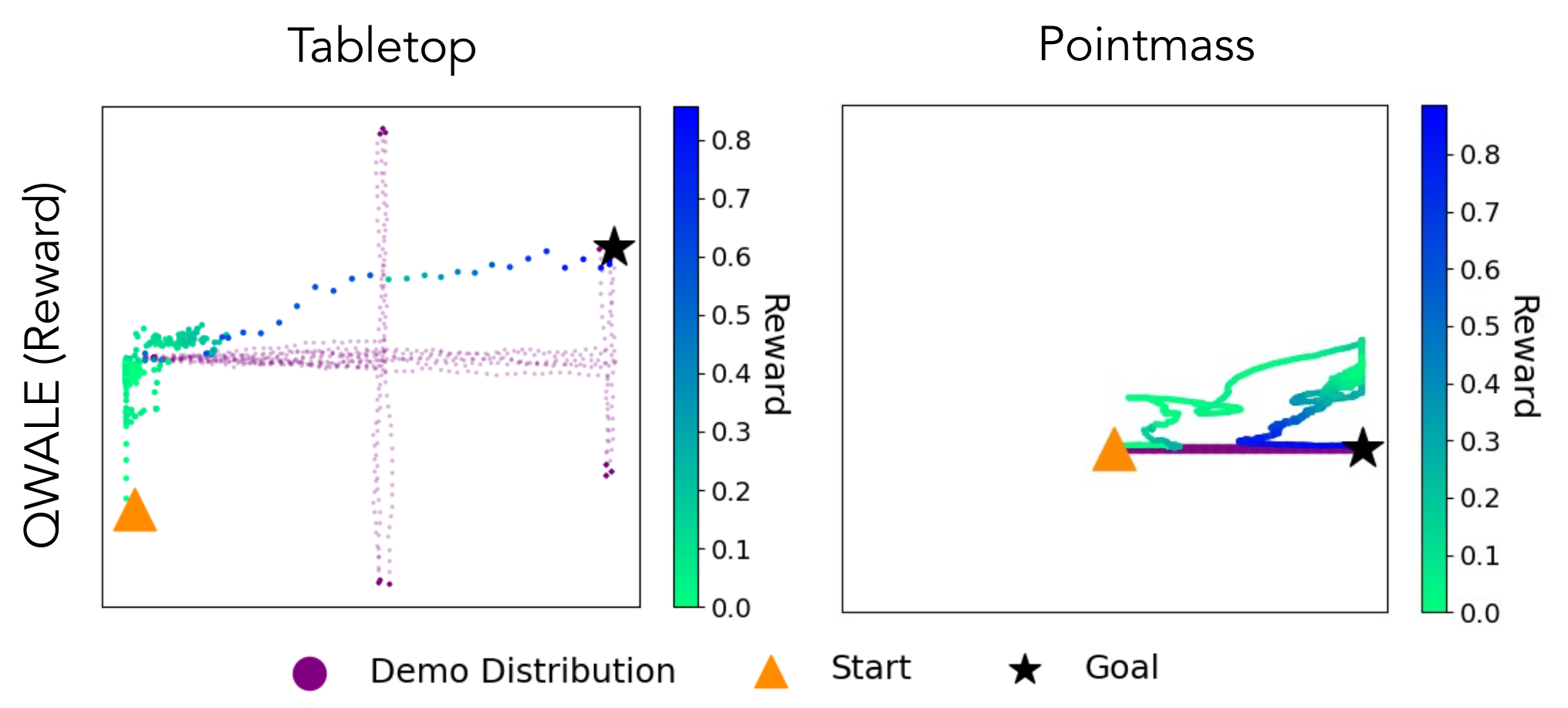}
    \caption{
        \small We visualize the online state visitation of the mug (w/ mirrored x-coordinate) in the Tabletop (left) and the agent in the Pointmass (right) environments of a single-life trial with \algo. The expert demo states are colored in purple, and we color the trajectories green to blue according to reward. The weighted discriminator allots higher reward to states closer to the goal.
    }
    \label{fig:visitation-reward}
\end{figure*}

Finally, to supplement Figure~\ref{fig:visitation}, we visualize the online state visitation of the mug in the Tabletop and the agent in the Pointmass environments of a single-life trial with \algo in Figure~\ref{fig:visitation-reward}. These plots show how \algo helps the agent recover from out-of-distribution states, as the reward gradually increases as the agent gets closer to the goal.



\end{document}